\pdfoutput=1

\documentclass[11pt]{article}

\usepackage{acl}

\usepackage{times}
\usepackage{latexsym}

\definecolor{}{rgb}{0.89, 0.44, 0.48}

\usepackage[T1]{fontenc}

\usepackage[utf8]{inputenc}

\usepackage{microtype}

\usepackage{multirow}
\usepackage{soul}
\usepackage{xcolor,colortbl}
\usepackage[shortlabels]{enumitem}
\usepackage{multirow}
\usepackage{array, makecell} 
\usepackage{subcaption}
\usepackage{bm}
\usepackage{mathrsfs} 
\usepackage{amsmath, amssymb}
\usepackage{cleveref}
\usepackage{todonotes}
\crefname{section}{§}{§§}
\Crefname{section}{§}{§§}
\Crefname{equation}{Eq.}{Eqs.}
\Crefname{figure}{Fig.}{Figs.}
\Crefname{tabular}{Tab.}{Tabs.}

\definecolor{Gray1}{rgb}{0.91,0.925, 0.937}
\definecolor{Gray2}{rgb}{0.87, 0.886, 0.902}
\definecolor{Gray3}{rgb}{0.808, 0.831, 0.855}
\definecolor{Gray4}{rgb}{0.678,0.71, 0.741}
\definecolor{Green0}{rgb}{0.909, 0.992, 0.886}
\definecolor{Green1}{rgb}{0.843, 0.960, 0.839}
\definecolor{Green2}{rgb}{0.635, 0.854, 0.627}
\definecolor{Red1}{rgb}{0.937, 0.686, 0.698}

\definecolor{darkgreen}{rgb}{0.0, 0.5, 0.0}
\definecolor{almond}{rgb}{0.99, 0.87, 0.9}
\definecolor{ghostwhite}{rgb}{0.97, 0.97, 1.0}
\definecolor{Blue1}{rgb}{0.792, 0.941, 0.973}
\definecolor{Blue2}{rgb}{0.678, 0.91, 0.957}
\definecolor{Blue3}{rgb}{0.565, 0.878, 0.937}
\definecolor{Blue4}{rgb}{0.282, 0.749, 0.89}

\definecolor{Yellow1}{rgb}{1, 0.914, 0.306}
\definecolor{Yellow2}{rgb}{1, 0.886, 0.275}
\definecolor{Yellow3}{rgb}{1, 0.855, 0.239}

\newcommand{\declarelogo}[0]{\includegraphics[height=.02\textwidth]{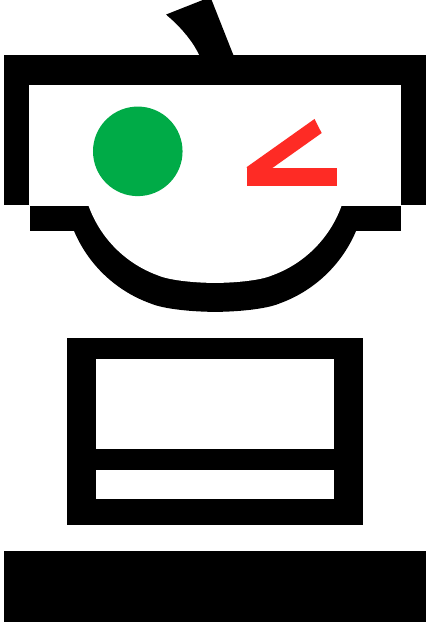}}
\newcommand{\nuslogo}[0]{\includegraphics[height=.02\textwidth]{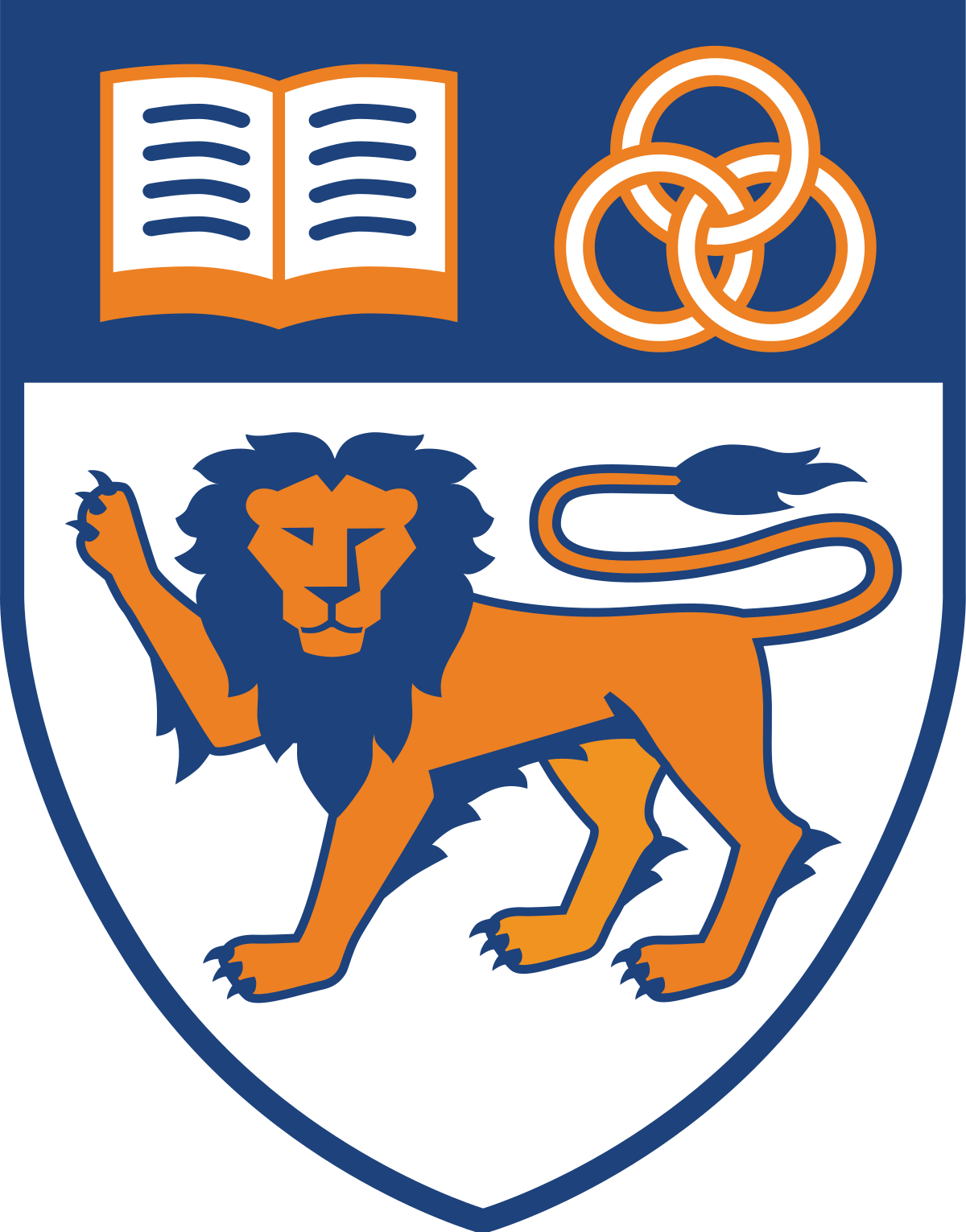}}

\title{Analyzing Modality Robustness in Multimodal Sentiment Analysis}

\author{Devamanyu Hazarika$^{\nuslogo}$\(^*\), Yingting Li\(^\alpha\) $^{\declarelogo}$\thanks{\ \ Equal Contribution.} ,\\ \textbf{Bo Cheng\(^\alpha\), \textbf{Shuai Zhao}\(^\alpha\), Roger Zimmermann$^{\nuslogo}$, Soujanya Poria}$^{\declarelogo}$\\
$^{\nuslogo}$ National University of Singapore, Singapore\\
\(^\alpha\) Beijing University of Posts and Telecommunications\\
$^{\declarelogo}$ DeCLaRe Lab, Singapore University of Technology and Design, Singapore \\
\fontsize{10}{10}\texttt{\{hazarika, rogerz\}@comp.nus.edu.sg}\\ 
\fontsize{10}{10}\texttt{\{cindyyting, chengbo, zhaoshuaiby\}@bupt.edu.cn}\\ 
\fontsize{10}{10}\texttt{sporia@sutd.edu.sg}\\ 
}

\usepackage{soul}
\begin{document}
\maketitle

\begin{abstract}
Building robust multimodal models are crucial for achieving reliable deployment in the wild. Despite its importance, less attention has been paid to identifying and improving the robustness of Multimodal Sentiment Analysis (MSA) models. In this work, we hope to address that by ($i$) Proposing simple diagnostic checks for modality robustness in a trained multimodal model. Using these checks, we find MSA models to be highly sensitive to a single modality, which creates issues in their robustness; ($ii$) We analyze well-known robust training strategies to alleviate the issues. Critically, we observe that robustness can be achieved \textit{without} compromising on the original performance. We hope our extensive study--performed across five models and two benchmark datasets--and proposed procedures would make robustness an integral component in MSA research. Our diagnostic checks and robust training solutions are simple to implement and available at \url{https://github.com/declare-lab/MSA-Robustness}
\end{abstract}

\section{Introduction}

Multimodal Sentiment Analysis (MSA) is a burgeoning field of research that has seen accelerated developments in recent years. Numerous models have been proposed that utilize multiple modalities such as audio, visual, and language signals to predict sentiments, emotions, and other forms of affect. While progress in MSA has been driven mainly by improvements in multimodal performance, we call for attention towards an equally important aspect in multimodal systems -- \textit{multimodal robustness}. Robustness is crucial when models are deployed in the wild, where it is common to encounter inadvertent errors in the source modalities due to data loss, data corruption, jitter, privacy issues, amongst others.

A well-known fact in the MSA research is that \textit{language} modality tends to be the most effective, which has prompted models to utilize language as its core modality~\cite{wu-etal-2021-text,DBLP:conf/icmi/HanCG0MP21,zeng2021making}. In this work, we focus on skewed dependence on language and try to understand how it affects the robustness of MSA models. Specifically, we ask,

\begin{figure}[t!]
    \centering
    \includegraphics[width=\linewidth]{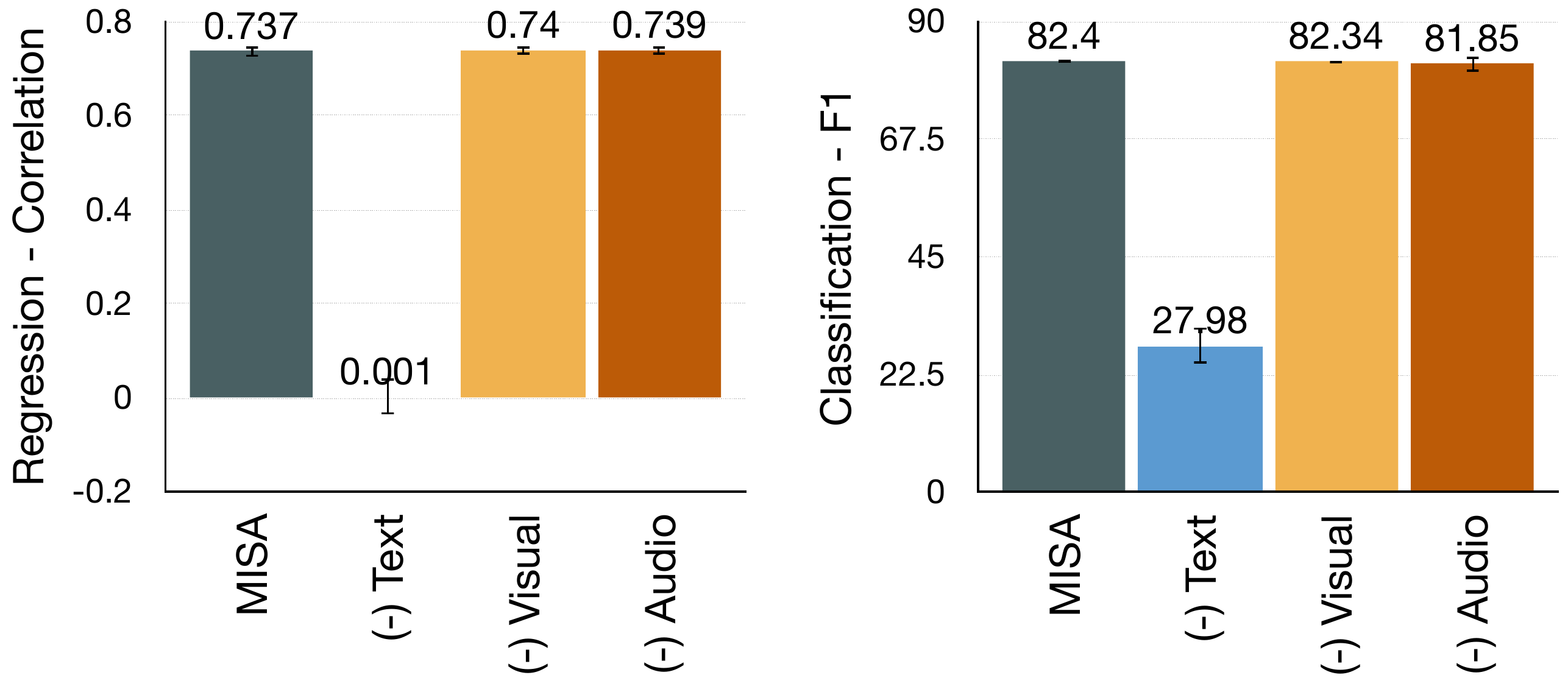}
    \caption{\footnotesize Removing modalities one at a time from the testing set of CMU-MOSI~\cite{7742221} on a trained MISA~\cite{DBLP:conf/mm/HazarikaZP20}.}
    \label{fig:motivation}
\end{figure}

\textbf{RQ1:} \textit{Are models in MSA over-reliant on a subset of modalities, particularly language?}

\textbf{RQ2:} \textit{If yes, what implications does it have on modality robustness?}

To answer \textbf{RQ1}, we look at \Cref{fig:motivation}. The figure illustrates a setup where we fully remove one modality during testing on the MISA model~\cite{DBLP:conf/mm/HazarikaZP20}. Here, we observe a sharp drop in performance when language modality is removed but do not see statistically significant drops when audio or visual modalities are removed. This observation aligns with recent findings in the MSA literature highlighting the dominance of language. 

This brings us to \textbf{RQ2} where we try to understand the robustness implications over this dominance. We design an elaborate study in \Cref{sec:testing_robustness}---over five state-of-the-art (SOTA) MSA models and across two benchmark datasets---where we propose diagnostic checks to understand \textit{modality robustness}, i.e., how robust are models against modality errors such as missing or noisy modalities.

Based on our findings, we then proceed to ask, 

\textbf{RQ3:} \textit{How can we improve the robustness of these models?}

\textbf{RQ4:} \textit{Does robust training lead to a performance trade-off?}

For \textbf{RQ3}, we study well-known robust training methods, that act as a pre-emptive strategy to reduce the performance drops. Critically, our training is \textit{model-agnostic} and can be easily included in any existing multimodal model (\Cref{sec:improving_robustness}). For \textbf{RQ4}, we observe that our method to improve robustness \textit{does not} trade-off with the final performance on the clean testing set, thus achieving similar performance as the original model.

\section{Related Works}
\label{sec:related_works}

While MSA has received increased attention in recent times, the topic of robustness has not taken center stage. Fortunately, few works have started changing this trend. \cite{DBLP:conf/nips/GatSSH20} reveals how multimodal classifiers often utilize a subset of modalities, which they addressed by inducing uniform contribution from all input modalities. In MSA, multiple works over-rely on language modality to improve the performance. \cite{wu-etal-2021-text} constructs a text-centered shared-private framework for multimodal fusion, and \cite{DBLP:conf/icmi/HanCG0MP21} obtains two text-related modal pairs and iteratively push the interaction between modalities to supplement information for better performance. While this has enabled performance boosts, our goal is to explore the double-edged nature of this feature and how it impacts robustness. Our motivation for diagnostics is similar to \cite{frank2021vision}, but unlike them, we do not perturb the raw data (such as image patches). Instead, we intervene on modality representations, which is easier to integrate with existing models and do not require prior knowledge of the modality structure. 

To address robustness in MSA, \cite{tsai2018learning} proposes a factored model that can accommodate modality drops. Also, \cite{ma2021smil} introduces modality drops during training and testing and uses meta-learning to make models robust. However, our work comprises some crucial distinctions: $i)$ Unlike these works, our diagnostics and robust training do not require sophisticated architecture and can be easily integrated into existing models. $ii)$ We perform an exhaustive analysis of robustness across multiple models, which is previously not done in the MSA literature.

\section{Testing Robustness via Diagnostic Checks}
\label{sec:testing_robustness}

In this section we perform an elaborate study on modality robustness by simulating potential issues with modality signals during testing (or deployment) of MSA models.

\begin{figure*}[t!]
    \centering
    \includegraphics[width=\linewidth]{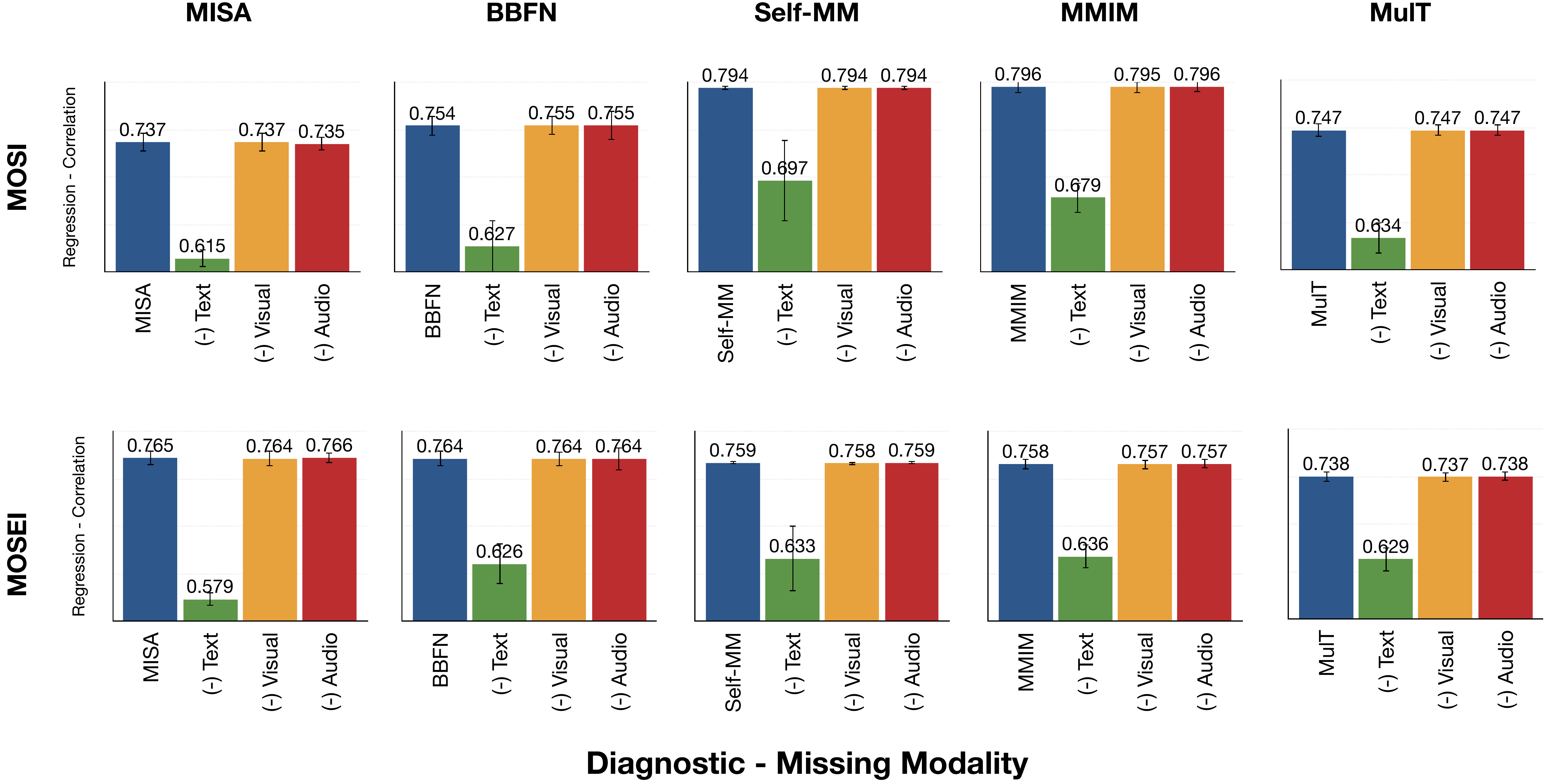}   
    \caption{\footnotesize Diagnostic checks (missing modality) for modality robustness in MOSI and MOSEI datasets. Results are averaged over three independent runs. Each modality error is applied to 30\% of testing data.}
    \label{fig:correlation_attack_missing}
\end{figure*}

\subsection{Experiment Setup}

\paragraph{Models.} In order to fully verify the universality of our experiments, we select a series of diverse SOTA models, ranging from RNN-based to Transformer-based architectures. These models work across different granularities from word-level to sentence-level variants:

$(i)$ \textbf{MISA}~\cite{DBLP:conf/mm/HazarikaZP20} is a popular model that generates modality-invariant and -specific features of multimodal data, to learn both shared and unique characteristics of each modality. $(ii)$ \textbf{BBFN}~\cite{DBLP:conf/icmi/HanCG0MP21} in a similar vein performs fusion and separation to increase cross-modal relevances and differences. This work acknowledges the dominance of text modality in MSA and proposes two text-centric bi-modal transformers to increase performance. $(iii)$ \textbf{Self-MM}~\cite{DBLP:conf/aaai/YuXYW21} focuses on the relationship between multi- and uni-modal predictions by multi-tasking consistencies and differences between them. $(iv)$ \textbf{MMIM}~\cite{DBLP:conf/emnlp/HanCP21} incorporates mutual information (MI) into MSA by maximizing MI at the input and fusion level. $(v)$ \textbf{MulT}~\cite{DBLP:conf/acl/TsaiBLKMS19} merges multimodal time series through multiple sets of directional pairwise cross-modal transformers. It accounts for long-range dependencies across modality elements to create a strong baseline (see~\Cref{sec:appendix_hyperparams}).

\paragraph{Datasets.} We consider two benchmark datasets widely used in the field of multimodal sentiment analysis, CMU-MOSI~\cite{7742221}, which is a popular dataset for studying the intensity of multimodal sentiment in the MSA field and CMU-MOSEI~\cite{bagher-zadeh-etal-2018-multimodal} which is a larger counterpart of MOSI with richer annotations and more diverse samples. Both these datasets contain short utterance videos and provide language, audio, and visual modality features.

\subsection{Proposed Diagnostic Checks}

We propose two diagnostic checks that introduce $i)$ \textit{Missing Modalities}, which drops (or nullifies) a modality from the input and $ii)$ \textit{Noisy Modalities} which include random changes to the modality representations, introduced via white Gaussian noise to the respective modality representations\footnote{While missing and noisy errors are predominant in the wild, we leave other potential forms of errors, such as affine transformations to the representations for future work.}. To simulate a realistic scenario, we apply these checks to 30\% of the testing data set\footnote{We set 30\% arbitrarily to simulate modality errors to a proportion of the input signals.}. Given the increased dependence on language modality in MSA models, we limit our study to errors introduced only in language modality without loss of generality. 

\begin{figure*}[t!]
    \centering
    \includegraphics[width=\linewidth]{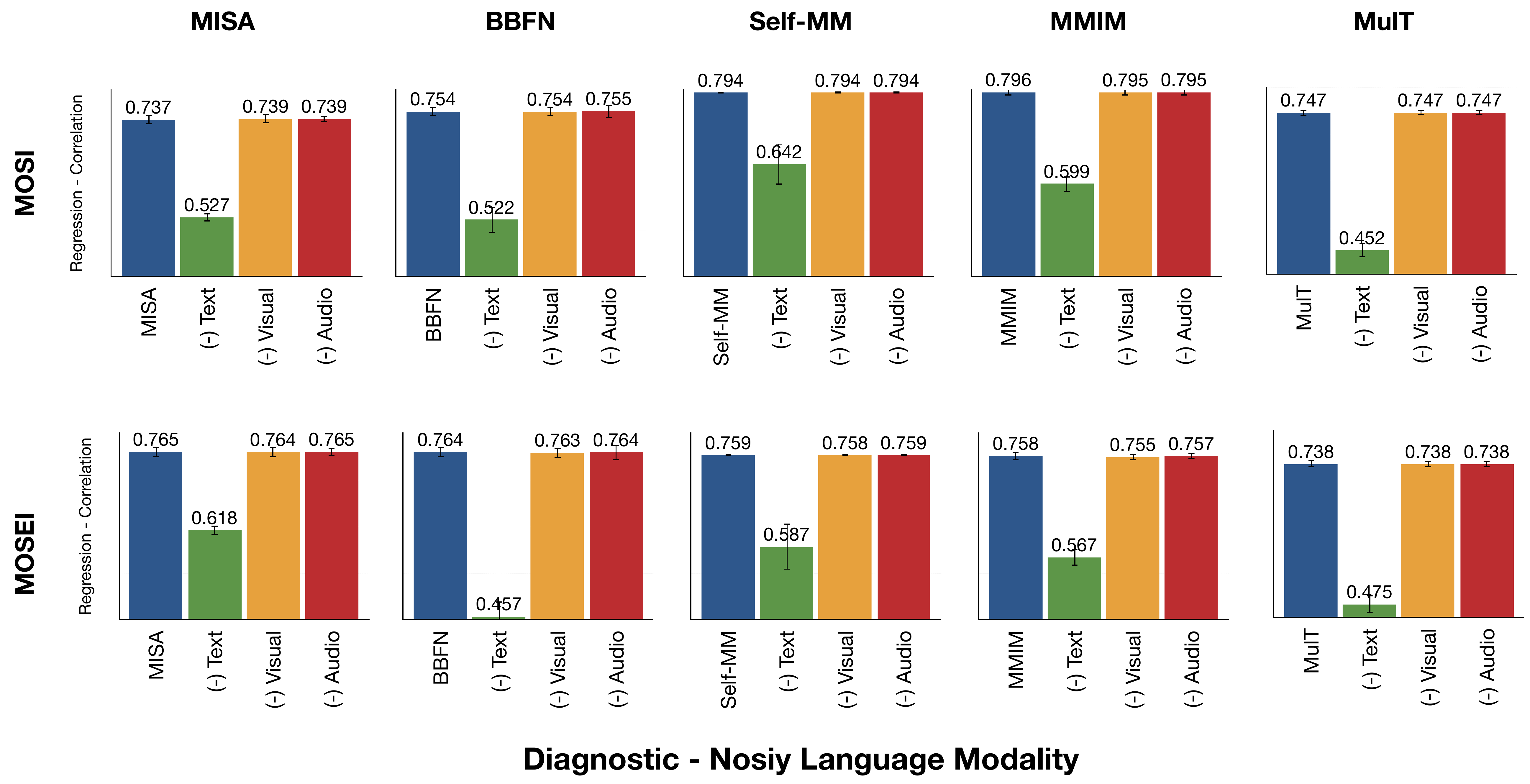}   
    \caption{\footnotesize Diagnostic checks (noisy modality) for modality robustness in MOSI and MOSEI datasets. Results are averaged over three independent runs. Each modality error is applied to 30\% of testing data.}
    \label{fig:correlation_attack_noise}
\end{figure*}

\paragraph{Procedure.}

We aim to intervene on modality representations to simulate modality errors. For the language modality $l$, all models map the sequence of tokens $U_{l}\in  R^{T_{l}}$ to its low-level embedding $\mathbf{U}_{l}\in  R^{T_{l} \times d_l}$ with $T_{l}$ tokens and $d_l$ embedding dimension. This low-level sequence is then encoded into hidden representations using an encoder of choice, such as BERT \cite{DBLP:conf/naacl/DevlinCLT19}, to achieve the language representation vector $\mathbf{u}_{l} = enc_{\theta _{l}}(U_{l}) \in \mathbb{R}^{d}$. We intervene on this representation and apply our diagnostics as follows.

We sample 30\% of $u_{l}$ from the testing set and modify them as $\mathbf{\hat{u}}_l = f(\mathbf{u}_l)$, where $f(\mathbf{x})$ is defined as either
$f(\mathbf{x}) = \mathbf{x} \odot \mathbf{0}$ for modality dropping (nulling the vector to 0s by element-wise multiplication) or $f(\mathbf{x}) = \mathbf{x} + \mathcal N(\mathbf{0}, \mathbf{1})$ to add white noise. The modified $\mathbf{\hat{u}}_l$ is then fed to the rest of the network as usual.




In the selected models, we apply diagnostics at different network locations. These include the representations before the hidden projection, such as in MISA, or fusion operation, such as in Self-MM. For MulT and BBFN, we apply the interventions right after the word embeddings. Detailed discussion on the location of interventions is provided in \Cref{sec:appendix_intervention}.


\paragraph{Observations.} \Cref{fig:correlation_attack_missing,fig:correlation_attack_noise} presents the results, where across both MOSI and MOSEI datasets, we find that language modality is highly sensitive to modality errors in the language source (across all models). This trend is observed for both missing and noisy modality checks, thus highlighting the concerns over robustness of these SOTA models. These diagnostic checks are easy to analyze, and we hope they will become an integrated part of the model-development pipeline in MSA. 

\section{Robust Training}
\label{sec:improving_robustness}

In this section, we explore how to reduce the sensitivity of the models to the dominant modality, i.e., language. One of the popular ways to alleviate such issues is to \textit{teach} the model such scenarios during training. We dub this approach as \textit{modality-perturbation}, which is conceptually similar to removing or masking modalities in~\cite{ma2021smil,georgiou2021m3} or adding noise in~\cite{miyato2018virtual}. It simulates the modality errors during training so that the model learns to expect such events during testing/deployment. The procedure is as follows,

\begin{enumerate}[leftmargin=*,itemsep=0pt]
    \item \textbf{Training:} 
    \begin{enumerate}[leftmargin=*,itemsep=0pt]
        \item For a particular batch of data, sample a proportion of the data to be perturbed.
        \item Similar to the diagnostic checks in \Cref{sec:testing_robustness}, perturb the dominant modality (in our case, language) of half of this data with \textit{missing} and the other half with \textit{noisy} perturbation. Repeat both these steps for the next batch. 
    \end{enumerate}  
    \item \textbf{Testing:} Apply the diagnostic checks as in \Cref{sec:testing_robustness}.
\end{enumerate}

This simple approach can be interpreted as regularization akin to dropouts or noising strategies used in de-noising auto-encoders.  

\subsection{Results}

\paragraph{Robustness.}

\Cref{tab:performance_drop} presents the results, where we perform \textit{balanced} perturbation between missing and noisy modalities. For the 30\% perturbable data in training, we drop the language modality on 15\% and for the other 15\%, we add noise. This setting improves the diagnostics in both kinds of errors. \Cref{sec:appendix_perturbation} presents results on other proportions of the training data.


With balanced perturbation, (BBFN-MOSI) reduces the relative drop on \textit{missing} language diagnostic by 31\% (in F1) and 98\% on \textit{noise}. Also, \textit{missing} drop reduces by 11\% in Corr and by 99\% for \textit{noisy} diagnostic. (Self-MM, MOSI) increases the relative drop in Corr slightly on \textit{missing} diagnostic, but in all other cases, it is significantly reduced. For example, the F1 drop on MOSI for \textit{noisy} diagnostic reduces significantly by 93\%. \Cref{tab:performance_drop} also shows that our method performs well on both RNN-based and Transformer-based models, demonstrating the wide applicability of our method.

\paragraph{Performance Trade-off.}

While alleviating robustness via regularization is well-known in the literature, there is often a trade-off with absolute performance in the original testing setup. Most approaches that achieve robustness take a hit at their best performance on clean input~\cite{DBLP:journals/corr/abs-2201-00464} ~\cite{DBLP:journals/corr/abs-1901-00532} ~\cite{DBLP:conf/eccv/SuZCYCG18} ~\cite{DBLP:conf/iclr/TsiprasSETM19}. This raises the question of whether introducing \textit{modality-perturbation} reduces the performance of the model on the original testing set. 

We find the answer to this is \textit{No}. Surprisingly, our robust training procedure \textit{does not} degrade in its original performance and can perform similar to the original model variants. This is highly ideal as we achieve robustness without compromising on performance in clean data.


\begin{table}[t]
    \centering
    \small
    \resizebox{\linewidth}{!}{
    \setcellgapes{2pt}
    \bgroup
    \def\arraystretch{1.2}%
    \begin{tabular}{ |l|l|l|r|r|r|r|  }
        \hline
         & \multirow{2}{*}{\makecell[l]{\textbf{Diagnostic}\\\textbf{ (30\%)}}} & \multirow{2}{*}{\makecell[l]{\textbf{Robust }\\\textbf{Training}}}&\multicolumn{2}{c|}{\textbf{MOSI}} & \multicolumn{2}{c|}{\textbf{MOSEI}}\\
         & & &  Corr & F1 &  Corr & F1\\
        \hline
        \hline
        \multirow{6}{*}{\rotatebox[origin=c]{90}{MISA}} 
        & \multirow{2}{*}{}                      & - & 0.737 & 82.40  & 0.765 & 85.76\\ 
        & & Yes & 0.736 & 81.42 & 0.767 & 85.97 \\ \cline{2-7}
        & \multirow{2}{*}{ missing}              & - & \textcolor{red}{$\downarrow$ 0.122} & \cellcolor{Red1!40} \textcolor{red}{$\downarrow$ 11.53} & \cellcolor{Red1!40} \textcolor{red}{$\downarrow$ 0.186} &  \cellcolor{Red1!40} \textcolor{red}{$\downarrow$ 8.45}\\ 
        & & Yes & \cellcolor{Red1!40} \textcolor{red}{$\downarrow$ 0.210}& \textcolor{red}{$\downarrow$ 9.96 } & \textcolor{red}{$\downarrow$ 0.147} & \textcolor{red}{$\downarrow$ 8.26}\\ \cline{2-7} 
        & \multirow{2}{*}{ noise}                & - & \textcolor{red}{$\downarrow$ 0.122} & \cellcolor{Red1!40} \textcolor{red}{$\downarrow$ 11.67} & \cellcolor{Red1!40} \textcolor{red}{$\downarrow$ 0.136} & \cellcolor{Red1!40} \textcolor{red}{$\downarrow$ 8.36}\\ 
        & & Yes & \cellcolor{Red1!40} \textcolor{red}{$\downarrow$ 0.163} & \textcolor{red}{$\downarrow$ 10.10} & \textcolor{red}{$\downarrow$ 0.002} & \textcolor{red}{$\downarrow$ 0.19}\\  
        \hline
        \hline
        \multirow{6}{*}{\rotatebox[origin=c]{90}{BBFN}} 
        &\multirow{2}{*}{}                      & -  & 0.754 & 83.12  & 0.764 & 85.70\\  
        & & Yes & 0.754 & 83.28  & 0.763 & 85.43\\  \cline{2-7}
        & \multirow{2}{*}{ missing}              & - & \cellcolor{Red1!40} \textcolor{red}{$\downarrow$ 0.127} & \cellcolor{Red1!40} \textcolor{red}{$\downarrow$ 10.55} & \cellcolor{Red1!40} \textcolor{red}{$\downarrow$ 0.139} & \cellcolor{Red1!40} \textcolor{red}{$\downarrow$ 10.57}\\ 
        & & Yes & \textcolor{red}{$\downarrow$ 0.119} & \textcolor{red}{$\downarrow$ 7.28}  & \textcolor{red}{$\downarrow$ 0.124} & \textcolor{red}{$\downarrow$ 7.88 }\\ \cline{2-7} \cline{2-7}
        & \multirow{2}{*}{ noise}                & - & \cellcolor{Red1!40} \textcolor{red}{$\downarrow$ 0.232} & \cellcolor{Red1!40} \textcolor{red}{$\downarrow$ 8.58}  & \cellcolor{Red1!40} \textcolor{red}{$\downarrow$ 0.308} & \cellcolor{Red1!40} \textcolor{red}{$\downarrow$ 9.62 }\\ 
        & & Yes & \textcolor{red}{$\downarrow$ 0.046} & \textcolor{red}{$\downarrow$ 0.16}  & \textcolor{red}{$\downarrow$ 0.003} & \textcolor{red}{$\downarrow$ 0.23 }\\
        \hline
        \hline
        \multirow{6}{*}{\rotatebox[origin=c]{90}{Self-MM}} 
        & \multirow{2}{*}{}                      & - &  0.794 & 85.61 & 0.759 & 84.62\\ 
        & & Yes &  0.790 & 84.73 & 0.754 & 84.67\\  \cline{2-7} \cline{2-7}
        & \multirow{2}{*}{ missing}              & - & \textcolor{red}{$\downarrow$ 0.099} & \cellcolor{Red1!40} \textcolor{red}{$\downarrow$ 11.74} & \cellcolor{Red1!40} \textcolor{red}{$\downarrow$ 0.126} & \cellcolor{Red1!40} \textcolor{red}{$\downarrow$ 9.04}\\ 
        & & Yes & \cellcolor{Red1!40} \textcolor{red}{$\downarrow$ 0.120} & \textcolor{red}{$\downarrow$ 9.66}  & \textcolor{red}{$\downarrow$ 0.122} & \textcolor{red}{$\downarrow$ 6.86}\\ \cline{2-7} \cline{2-7}
        & \multirow{2}{*}{ noise}                & - & \cellcolor{Red1!40} \textcolor{red}{$\downarrow$ 0.154} & \cellcolor{Red1!40} \textcolor{red}{$\downarrow$ 8.35}  & \cellcolor{Red1!40} \textcolor{red}{$\downarrow$ 0.172} & \cellcolor{Red1!40} \textcolor{red}{$\downarrow$ 9.48}\\ 
        & & Yes & \textcolor{red}{$\downarrow$ 0.041} & \textcolor{red}{$\downarrow$ 0.58}  & \textcolor{red}{$\downarrow$ 0.051} & \textcolor{red}{$\downarrow$ 1.02}\\
        \hline
        \hline
        \multirow{6}{*}{\rotatebox[origin=c]{90}{MMIM}} 
        & \multirow{2}{*}{}                      & - & 0.796 & 86.02 & 0.758 & 84.89\\  
        & & Yes & 0.784 & 84.67 & 0.751 & 83.15\\  \cline{2-7}
        & \multirow{2}{*}{ missing}              & - & \textcolor{red}{$\downarrow$ 0.117} & \textcolor{red}{$\downarrow$ 9.37} & \cellcolor{Red1!40} \textcolor{red}{$\downarrow$ 0.122} & \cellcolor{Red1!40} \textcolor{red}{$\downarrow$ 8.15}\\ 
        & & Yes                                      & \cellcolor{Red1!40} \textcolor{red}{$\downarrow$ 0.146} & \cellcolor{Red1!40} \textcolor{red}{$\downarrow$ 10.48} & \textcolor{red}{$\downarrow$ 0.115} & \textcolor{red}{$\downarrow$  6.62} \\ \cline{2-7}
        & \multirow{2}{*}{ noise}                & - & \cellcolor{Red1!40} \textcolor{red}{$\downarrow$ 0.197} & \cellcolor{Red1!40} \textcolor{red}{$\downarrow$ 9.55} & \cellcolor{Red1!40} \textcolor{red}{$\downarrow$ 0.191} & \cellcolor{Red1!40} \textcolor{red}{$\downarrow$ 9.18}\\ 
        & & Yes                                      & \textcolor{red}{$\downarrow$ 0.180} & \textcolor{red}{$\downarrow$ 8.88}  & \textcolor{red}{$\downarrow$ 0.096} & \textcolor{red}{$\downarrow$  4.41} \\ 
        \hline
        \hline
        \multirow{6}{*}{\rotatebox[origin=c]{90}{Mult}} 
        & \multirow{2}{*}{}                      & - & 0.747 & 82.25 & 0.738 & 83.37\\  
        & & Yes                                      &  0.744 & 82.21 & 0.745 & 83.95\\  \cline{2-7}
        & \multirow{2}{*}{ missing}              & - & \textcolor{red}{$\downarrow$ 0.113} & \cellcolor{Red1!40} \textcolor{red}{$\downarrow$ 12.17} & \textcolor{red}{$\downarrow$ 0.109} & \textcolor{red}{$\downarrow$  6.60}    \\ 
        & & Yes                                      & \cellcolor{Red1!40} \textcolor{red}{$\downarrow$ 0.117} & \textcolor{red}{$\downarrow$ 9.63 } & \cellcolor{Red1!40} \textcolor{red}{$\downarrow$ 0.113} & \cellcolor{Red1!40} \textcolor{red}{$\downarrow$  7.03}    \\ \cline{2-7}
        & \multirow{2}{*}{ noise}                & - & \cellcolor{Red1!40} \textcolor{red}{$\downarrow$ 0.295} & \cellcolor{Red1!40} \textcolor{red}{$\downarrow$ 8.99 } & \cellcolor{Red1!40} \textcolor{red}{$\downarrow$ 0.263} & \cellcolor{Red1!40} \textcolor{red}{$\downarrow$  7.95}    \\ 
        & & Yes                                      & \textcolor{red}{$\downarrow$ 0.068} & \textcolor{red}{$\downarrow$ 6.13 } & \textcolor{red}{$\downarrow$ 0.001} & \textcolor{darkgreen}{$\uparrow$ 0.02} \\ 
        \hline
    \end{tabular}
    \egroup
    }
    \caption{Robust Training is performed with 15\% missing and 15\% noise perturbation. Results are averaged over 3 random runs. More perturbation results are provided in \Cref{sec:appendix_perturbation}. Higher drops between Non-robust and robust training (consecutive rows) are highlighted.}
    \label{tab:performance_drop}
\end{table}

\section{Conclusion}

In this work, we performed a systematic study that demonstrate the double-edged nature of dominant modality in SOTA MSA models. Our analysis using diagnostic checks reveal high susceptibility to performance drops when presented with unwanted errors in their representations. 

To alleviate the issues, we also study robust training methods that uses modality perturbations. Critically, we find that robustness and performance can co-exist without an explicit trade-off. These improvements demonstrate a positive nudge in the effort to achieve robustness and we believe there remains significant room for improvement. With this work, by proposing simple and easy-to-integrate diagnostic checks and training methods, we hope to permeate discussions on robustness into mainstream MSA research.

\section*{Acknowledgements}

This research is supported by $(i)$ T2MOE2008 awarded by Singapore’s MoE under its Tier-2 grant scheme, and $(ii)$ SUTD SRG grant \#T1SRIS19149. We also thank the anonymous reviewers for their valuable comments.

\bibliography{references}
\bibliographystyle{acl_natbib}

\newpage
\appendix

\section{Model Details}
\label{sec:appendix_intervention}

\paragraph{MISA:} We get the MISA model from its official repository \footnote{https://github.com/declare-lab/MISA/tree/ec42faddde0d210cf7368aebf2118fe9570e7102}. In this model, we apply the interventions at the following encoded language representation from the original paper: 

\begin{align}
\mathbf{u}_{l}=\operatorname{Bert}\left(\mathbf{U}_{l} ; \theta_{l}^{bert}\right) \label{eqn:misa1} \\
\mathbf{\hat{u}}_{l}= f(\mathbf{\hat{u}}_{l}) \\
\mathbf{h}_{l}^{c}=E_{c}\left(\mathbf{\hat{u}}_{l} ; \theta^{c}\right), \quad \mathbf{h}_{l}^{p}=E_{p}\left(\mathbf{\hat{u}}_{l} ; \theta_{l}^{p}\right) \label{eqn:misa2}
\end{align}

That is, the interventions are applied before the language representation is projected to its shared and private subspaces.

\paragraph{BBFN:}We get the BBFN model from its official repository \footnote{https://github.com/declare-lab/BBFN/tree/be15f947ed7539b3c54381e453f09439466ed915}. In this model, we execute the interventions after the following language embedding from the original paper.

\begin{align}
    \mathbf{M}_{l}=\left(m_{0}, m_{1}, \ldots, m_{n+1}\right) \label{eqn:bbfn1} \\
    \mathbf{\hat{M}}_{l}= f(\mathbf{M}_{l}) \label{eqn:bbfn2}
\end{align}

\paragraph{Self-MM:}We get the Self-MM model from its official repository\footnote{https://github.com/thuiar/Self-MM}. In this model, we set the interventions after the language features encoded below from the original paper.

\begin{align}
    \mathbf{F}_{l}=B E R T\left(I_{l} ; \theta_{l}^{b e r t}\right) \in R^{d_{l}} \label{eqn:self_mm1} \\
    \mathbf{\hat{F}}_{l}= f(\mathbf{F}_{l}) \label{eqn:self_mm2}
\end{align}

\paragraph{MMIM:}We get the MMIM model from its official repository \footnote{https://github.com/declare-lab/Multimodal-Infomax}. For this model, we perform the interventions after the following encoded language representation from the original paper.

\begin{align}
    \mathbf{x}_{l}=\operatorname{BERT}\left(X_{l} ; \theta_{l}^{B E R T}\right) \label{eqn:mmim1} \\
    \mathbf{\hat{x}}_{l}= f(\mathbf{x}_{l}) \label{eqn:mmim2}
\end{align}

\paragraph{MulT:}We get the MulT model from its official repository \footnote{https://github.com/yaohungt/Multimodal-Transformer}. We intervene in this model after the following encoded language representation:.

\begin{align}
\mathbf{x}_{l}=\operatorname{Conv1D}(X_{l}, k_{l}) \in \mathbb{R}^{T_{l} \times d} \label{eqn:mult1}  \\
\mathbf{\hat{x}}_{l}= f(\mathbf{x}_{l}) \label{eqn:mult2}
\end{align}

\section{Reproducing Results}
\label{sec:appendix_hyperparams}

\begin{table}[t!]
    \centering\renewcommand\cellalign{lc}
    \setcellgapes{3pt}\makegapedcells
    \resizebox{\linewidth}{!}{
    \begin{tabular}{ |c|l|c|c|  }
    \hline
    Models & Item & CMU-MOSI & CMU-MOSEI\\
    \cline{1-4}
    \multirow{3}{*}{MISA} & Learning rate & 1e-5 & 4e-5 \\
    & Optimizer & RMSprop & Adam \\
    & Activation & hardtanh & relu \\
    \hline
    \hline
    \multirow{1}{*}{BBFN} & Learning rate & 1e-4 & 5e-5 \\
    \hline
    \hline
    \multirow{7}{*}{MMIM} & Batch size & 32 & 64 \\
    & learning rate $\eta_{lld}$ & 4e-3 & 1e-3 \\
    & learning rate $\eta_{main}$ & 1e-3 & 5e-4 \\
    & $\alpha$ & 0.3 & 0.1 \\
    & $\beta$ & 0.1 & 0.05 \\
    & V-LSTM hidden dim & 32 & 64 \\
    & A-LSTM hidden dim & 32 & 16 \\
    \hline
    \hline
    \multirow{12}{*}{MulT} & Batch size & 128 & 16 \\
    & Learning rate & 1e-4 & 1e-4 \\
    & Optimizer & Adam & Adam \\
    & \makecell[l]{Transformers Hidden \\ Unit Size d} & 30 & 30 \\
    & \makecell[l]{No. of Crossmodal \\Attention Heads} & 10 & 10 \\
    & \makecell[l]{No. of Crossmodal\\a Blocks D} & 4 & 4 \\
    & \makecell[l]{Textual Embedding \\ Dropout} & 0.3 & 0.3 \\
    & \makecell[l]{Crossmodal Attention \\ Block Dropout} & 0.2 & 0.1 \\
    & Output Dropout & 0.2 & 0.1 \\
    & Gradient Clip & 0.8 & 1.0 \\
    & No. of Epochs & 100 & 20 \\
    & Use Bert & Yes & Yes \\
    \hline
    \end{tabular}
    }
    \caption{Hyper-parameter config used to train  the models.}
    \label{tab:hyperparams}
\end{table}

For each model we train the models to achieve performances close to reported in the respective papers. \Cref{tab:hyperparams} presents the hyper-parameters we used to reproduce their results.

\section{Additional Results on Modality-Perturbation}
\label{sec:appendix_perturbation}

We also analyze with varying proportions of perturbations in the training and testing phase, respectively. As seen in \Cref{tab: misa_robust}, as the noise gradually increases from 5\% to 15\%, the drop of Corr in MOSI is gradually reduced , which shows the robustness is getting better, until it reaches the optimum at 30\% perturbation (15\% missing + 15\% noise). In other models, 30\% perturbation is also advantageous. For example, in \Cref{tab: mult_robust}, (Mult, MOSEI) reduces Corr drop while improving F1 performance in 30\% perturbation. Although it is only a small improvement at present, we believe that there will be more meaningful improvements in the future.

\begin{table*}[t]
    \centering
     \small
     \resizebox{0.7\textwidth}{!}{
    \begin{tabular}{ |l|l|l|r|r|r|r|  }
        \hline
         &  \multirow{2}{*}{\makecell[l]{\textbf{Robust }\\\textbf{Training}}}& \multirow{2}{*}{ \textbf{Diagnostic} }&\multicolumn{2}{c|}{\textbf{MOSI}} & \multicolumn{2}{c|}{\textbf{MOSEI}}\\
         & & &  Corr & F1 &  Corr & F1\\
        \hline
        \hline
        \multirow{36}{*}{\rotatebox[origin=c]{90}{MISA}} &\multirow{9}{*}{-} &   & 0.737 & 82.40  & 0.765 & 85.76\\  \cline{3-7}
        && missing 5\%  & \textcolor{red}{$\downarrow$ 0.011 }   & \textcolor{red}{$\downarrow$ 1.18 }  & \textcolor{red}{$\downarrow$  0.042 } & \textcolor{red}{$\downarrow$ 1.57 }\\
        && noise   5\%  & \textcolor{red}{$\downarrow$ 0.022 }   & \textcolor{red}{$\downarrow$ 0.75 }  &          $\rightarrow$  0             &          $\rightarrow$  0          \\
        && missing 10\% & \textcolor{red}{$\downarrow$ 0.031 }   & \textcolor{red}{$\downarrow$ 2.19 }  & \textcolor{red}{$\downarrow$  0.073 } & \textcolor{red}{$\downarrow$ 2.41 }\\
        && noise   10\% & \textcolor{red}{$\downarrow$ 0.047 }   & \textcolor{red}{$\downarrow$ 2.56 }  & \textcolor{red}{$\downarrow$  0.031 } & \textcolor{red}{$\downarrow$ 1.75 }\\
        && missing 15\% & \textcolor{red}{$\downarrow$ 0.050 }   & \textcolor{red}{$\downarrow$ 3.99 }  & \textcolor{red}{$\downarrow$  0.126 } & \textcolor{red}{$\downarrow$ 4.10 }\\
        && noise   15\% & \textcolor{red}{$\downarrow$ 0.076 }   & \textcolor{red}{$\downarrow$ 4.05 }  & \textcolor{red}{$\downarrow$  0.068 } & \textcolor{red}{$\downarrow$ 3.88 }\\
        && missing 30\% & \textcolor{red}{$\downarrow$ 0.122 }   & \textcolor{red}{$\downarrow$ 11.53}  & \textcolor{red}{$\downarrow$  0.186 } & \textcolor{red}{$\downarrow$ 8.45 }\\
        && noise   30\% & \textcolor{red}{$\downarrow$ 0.210 }   & \textcolor{red}{$\downarrow$ 9.96 }  & \textcolor{red}{$\downarrow$  0.147 } & \textcolor{red}{$\downarrow$ 8.26 }\\
        \cline{2-7}
        \cline{2-7}
        &\multirow{9}{*}{\makecell[l]{5\% missing \\ 5\% noise}} &    & 0.714 & 80.99 & 0.765 & 85.68 \\  \cline{3-7}
        && missing 5\%  & \textcolor{red}{$\downarrow$ 0.010 }   & \textcolor{red}{$\downarrow$ 1.38 }  & \textcolor{red}{$\downarrow$  0.042 } & \textcolor{red}{$\downarrow$ 1.57 }\\
        && noise   5\%  & \textcolor{red}{$\downarrow$ 0.032 }   & \textcolor{red}{$\downarrow$ 1.37 }  &          $\rightarrow$  0             & $\rightarrow$ 0 \\
        && missing 10\% & \textcolor{red}{$\downarrow$ 0.030 }   & \textcolor{red}{$\downarrow$ 2.17 }  & \textcolor{red}{$\downarrow$  0.074 } & \textcolor{red}{$\downarrow$ 2.41 }\\
        && noise   10\% & \textcolor{red}{$\downarrow$ 0.055 }   & \textcolor{red}{$\downarrow$ 2.44 }  & \textcolor{red}{$\downarrow$  0.041 } & \textcolor{red}{$\downarrow$ 2.22 }\\
        && missing 15\% & \textcolor{red}{$\downarrow$ 0.044 }   & \textcolor{red}{$\downarrow$ 4.52 }  & \textcolor{red}{$\downarrow$  0.126 } & \textcolor{red}{$\downarrow$ 0.41 }\\
        && noise   15\% & \textcolor{red}{$\downarrow$ 0.094 }   & \textcolor{red}{$\downarrow$ 3.65 }  & \textcolor{red}{$\downarrow$  0.058 } & \textcolor{red}{$\downarrow$ 3.53 }\\
        && missing 30\% & \textcolor{red}{$\downarrow$ 0.121 }   & \textcolor{red}{$\downarrow$ 11.08}  & \textcolor{red}{$\downarrow$  0.186 } & \textcolor{red}{$\downarrow$ 8.37 }\\
        && noise   30\% & \textcolor{red}{$\downarrow$ 0.201 }   & \textcolor{red}{$\downarrow$ 8.71 }  & \textcolor{red}{$\downarrow$  0.150 } & \textcolor{red}{$\downarrow$ 8.12 }\\
        \cline{2-7}
        \cline{2-7}
        &\multirow{9}{*}{\makecell[l]{10\% missing \\ 10\% noise}} &    & 0.734 & 81.40 & 0.765 & 85.68 \\  \cline{3-7}
        && missing 5\%  & \textcolor{red}{$\downarrow$ 0.011 }   & \textcolor{red}{$\downarrow$ 1.04 }  & \textcolor{red}{$\downarrow$  0.022 } & \textcolor{red}{$\downarrow$ 1.58 }\\
        && noise   5\%  & \textcolor{red}{$\downarrow$ 0.028 }   & \textcolor{red}{$\downarrow$ 1.20 }  &          $\rightarrow$  0             & $\rightarrow$ 0 \\
        && missing 10\% & \textcolor{red}{$\downarrow$ 0.030 }   & \textcolor{red}{$\downarrow$ 2.05 }  & \textcolor{red}{$\downarrow$  0.042 } & \textcolor{red}{$\downarrow$ 2.54 }\\
        && noise   10\% & \textcolor{red}{$\downarrow$ 0.045 }   & \textcolor{red}{$\downarrow$ 2.27 }  & \textcolor{red}{$\downarrow$  0.001 } & \textcolor{red}{$\downarrow$ 0.02 }\\
        && missing 15\% & \textcolor{red}{$\downarrow$ 0.049 }   & \textcolor{red}{$\downarrow$ 3.54 }  & \textcolor{red}{$\downarrow$  0.071 } & \textcolor{red}{$\downarrow$ 4.04 }\\
        && noise   15\% & \textcolor{red}{$\downarrow$ 0.073 }   & \textcolor{red}{$\downarrow$ 3.42 }  & \textcolor{red}{$\downarrow$  0.001 } & \textcolor{red}{$\downarrow$ 0.43 }\\
        && missing 35\% & \textcolor{red}{$\downarrow$ 0.118 }   & \textcolor{red}{$\downarrow$ 11.11}  & \textcolor{red}{$\downarrow$  0.135 } & \textcolor{red}{$\downarrow$ 8.17 } \\
        && noise   30\% & \textcolor{red}{$\downarrow$ 0.181 }   & \textcolor{red}{$\downarrow$ 9.55 }  & \textcolor{red}{$\downarrow$  0.035 } & \textcolor{red}{$\downarrow$ 0.54 }\\
        \cline{2-7}
        \cline{2-7}
        &\multirow{9}{*}{\makecell[l]{15\% missing \\ 15\% noise} } &   & 0.736 & 81.42 & 0.767 & 85.97 \\  \cline{3-7}
        && missing 5\%  & \textcolor{red}{$\downarrow$  0.012 } & \textcolor{red}{$\downarrow$ 1.50 } & \textcolor{red}{$\downarrow$ 0.021 }   & \textcolor{red}{$\downarrow$ 1.46 }  \\
        && noise   5\%  & \textcolor{red}{$\downarrow$  0.027 } & \textcolor{red}{$\downarrow$ 1.50 } &          $\rightarrow$  0              &          $\rightarrow$  0            \\
        && missing 10\% & \textcolor{red}{$\downarrow$  0.033 } & \textcolor{red}{$\downarrow$ 2.71 } & \textcolor{red}{$\downarrow$ 0.041 }   & \textcolor{red}{$\downarrow$ 2.43 }  \\
        && noise   10\% & \textcolor{red}{$\downarrow$  0.032 } & \textcolor{red}{$\downarrow$ 2.12 } & \textcolor{red}{$\downarrow$ 0.001 }   & \textcolor{red}{$\downarrow$ 0.03 }  \\
        && missing 15\% & \textcolor{red}{$\downarrow$  0.052 } & \textcolor{red}{$\downarrow$ 4.40 } & \textcolor{red}{$\downarrow$ 0.021 }   & \textcolor{red}{$\downarrow$ 4.11 }  \\
        && noise   15\% & \textcolor{red}{$\downarrow$  0.073 } & \textcolor{red}{$\downarrow$ 4.99 } & \textcolor{red}{$\downarrow$ 0.001 }   & \textcolor{red}{$\downarrow$ 0.12 }  \\
        && missing 30\% & \textcolor{red}{$\downarrow$  0.122 } & \textcolor{red}{$\downarrow$ 11.67} & \textcolor{red}{$\downarrow$ 0.136 }   & \textcolor{red}{$\downarrow$ 8.36 }  \\ 
        && noise   30\% & \textcolor{red}{$\downarrow$  0.163 } & \textcolor{red}{$\downarrow$ 10.10} & \textcolor{red}{$\downarrow$ 0.002 }   & \textcolor{red}{$\downarrow$ 0.19 }  \\
        \hline
    \end{tabular}
     }
    \caption{MISA Robust Training. Results are averaged over three random runs.}
    \label{tab: misa_robust}
\end{table*}
        
\begin{table*}[t]
    \centering
     \small
     \resizebox{0.7\textwidth}{!}{
    \begin{tabular}{ |l|l|l|r|r|r|r|  }
        \hline
         &  \multirow{2}{*}{\makecell[l]{\textbf{Robust }\\\textbf{Training}}}& \multirow{2}{*}{ \textbf{Diagnostic} }&\multicolumn{2}{c|}{\textbf{MOSI}} & \multicolumn{2}{c|}{\textbf{MOSEI}}\\
         & & &  Corr & F1 &  Corr & F1\\
        \hline
        \hline
        \multirow{36}{*}{\rotatebox[origin=c]{90}{BBFN}} &\multirow{9}{*}{-} &   & 0.754 & 83.12  & 0.764 & 85.70\\  \cline{3-7}
        && missing 5\%  & \textcolor{red}{$\downarrow$ 0.013 }   & \textcolor{red}{$\downarrow$ 1.51 }  & \textcolor{red}{$\downarrow$  0.020 } & \textcolor{red}{$\downarrow$ 1.97 }\\
        && noise   5\%  & \textcolor{red}{$\downarrow$ 0.034 }   & \textcolor{red}{$\downarrow$ 0.75 }  & \textcolor{red}{$\downarrow$  0.050 } & \textcolor{red}{$\downarrow$ 1.21 }\\
        && missing 10\% & \textcolor{red}{$\downarrow$ 0.028 }   & \textcolor{red}{$\downarrow$ 2.59 }  & \textcolor{red}{$\downarrow$  0.038 } & \textcolor{red}{$\downarrow$ 3.34 }\\
        && noise   10\% & \textcolor{red}{$\downarrow$ 0.093 }   & \textcolor{red}{$\downarrow$ 2.90 }  & \textcolor{red}{$\downarrow$  0.121 } & \textcolor{red}{$\downarrow$ 3.54 }\\
        && missing 15\% & \textcolor{red}{$\downarrow$ 0.032 }   & \textcolor{red}{$\downarrow$ 3.79 }  & \textcolor{red}{$\downarrow$  0.055 } & \textcolor{red}{$\downarrow$ 5.45 }\\
        && noise   15\% & \textcolor{red}{$\downarrow$ 0.080 }   & \textcolor{red}{$\downarrow$ 2.28 }  & \textcolor{red}{$\downarrow$  0.154 } & \textcolor{red}{$\downarrow$ 4.18 }\\
        && missing 30\% & \textcolor{red}{$\downarrow$ 0.127 }   & \textcolor{red}{$\downarrow$ 10.55 } & \textcolor{red}{$\downarrow$  0.139 } & \textcolor{red}{$\downarrow$ 10.57}\\
        && noise   30\% & \textcolor{red}{$\downarrow$ 0.232 }   & \textcolor{red}{$\downarrow$ 8.58 }  & \textcolor{red}{$\downarrow$  0.308 } & \textcolor{red}{$\downarrow$ 9.62}\\
        \cline{2-7}
        \cline{2-7}
        &\multirow{9}{*}{\makecell[l]{5\% missing \\ 5\% noise}} &    & 0.743 & 82.39 & 0.765 & 85.48 \\  \cline{3-7}
        && missing 5\%  & \textcolor{red}{$\downarrow$ 0.020 }  & \textcolor{red}{$\downarrow$ 0.94 }  & \textcolor{red}{$\downarrow$  0.017 } & \textcolor{red}{$\downarrow$ 1.00 }\\
        && noise   5\%  & \textcolor{red}{$\downarrow$ 0.002 }  &              $\rightarrow$ 0         &              $\rightarrow$ 0          & \textcolor{red}{$\downarrow$ 0.08 }\\
        && missing 10\% & \textcolor{red}{$\downarrow$ 0.039 }  & \textcolor{red}{$\downarrow$ 2.17 }  & \textcolor{red}{$\downarrow$  0.032 } & \textcolor{red}{$\downarrow$ 1.95 }\\
        && noise   10\% &             $\rightarrow$ 0           &             $\rightarrow$ 0          &              $\rightarrow$ 0          & \textcolor{red}{$\downarrow$ 0.06 }\\
        && missing 15\% & \textcolor{red}{$\downarrow$ 0.045 }  & \textcolor{red}{$\downarrow$ 2.18 }  & \textcolor{red}{$\downarrow$  0.050 } & \textcolor{red}{$\downarrow$ 3.57 }\\
        && noise   15\% & \textcolor{red}{$\downarrow$ 0.002 }  & \textcolor{red}{$\downarrow$ 0.15 }  & \textcolor{red}{$\downarrow$  0.001 } &             $\rightarrow$ 0        \\
        && missing 30\% & \textcolor{red}{$\downarrow$ 0.049 }  & \textcolor{red}{$\downarrow$ 1.92 }  & \textcolor{red}{$\downarrow$  0.122 } & \textcolor{red}{$\downarrow$ 7.60 }\\
        && noise   30\% &             $\rightarrow$ 0           & \textcolor{red}{$\downarrow$ 0.20 }  & \textcolor{red}{$\downarrow$  0.001 } & \textcolor{red}{$\downarrow$ 0.04 }\\
        \cline{2-7}
        \cline{2-7}
        &\multirow{9}{*}{\makecell[l]{10\% missing \\ 10\% noise}} &   & 0.742 & 81.66 & 0752 & 85.15 \\  \cline{3-7}
        
        && missing  5\% & \textcolor{red}{$\downarrow$ 0.018 }  & \textcolor{red}{$\downarrow$ 1.66 }  & \textcolor{red}{$\downarrow$  0.018} & \textcolor{red}{$\downarrow$ 1.62 }\\
        && noise    5\% & \textcolor{red}{$\downarrow$ 0.001 }  &               $\rightarrow$ 0        & \textcolor{red}{$\downarrow$  0.022} & \textcolor{red}{$\downarrow$ 0.05 }\\
        && missing 10\% & \textcolor{red}{$\downarrow$ 0.034 }  & \textcolor{red}{$\downarrow$ 2.87 }  & \textcolor{red}{$\downarrow$  0.035} & \textcolor{red}{$\downarrow$ 2.87 }\\
        && noise   10\% & \textcolor{red}{$\downarrow$ 0.001 }  &               $\rightarrow$ 0        & \textcolor{red}{$\downarrow$  0.003} & \textcolor{red}{$\downarrow$ 0.09 }\\
        && missing 15\% & \textcolor{red}{$\downarrow$ 0.036 }  & \textcolor{red}{$\downarrow$ 3.63 }  & \textcolor{red}{$\downarrow$  0.050} & \textcolor{red}{$\downarrow$ 4.47 }\\
        && noise   15\% & \textcolor{red}{$\downarrow$ 0.001 }  & \textcolor{red}{$\downarrow$ 0.01 }  & \textcolor{red}{$\downarrow$  0.004} & \textcolor{red}{$\downarrow$ 0.07 }\\
        && missing 30\% & \textcolor{red}{$\downarrow$ 0.126 }  & \textcolor{red}{$\downarrow$ 9.75 }  & \textcolor{red}{$\downarrow$  0.125} & \textcolor{red}{$\downarrow$ 9.40 }\\
        && noise   30\% & \textcolor{red}{$\downarrow$ 0.003 }  & \textcolor{red}{$\downarrow$ 0.46 }  & \textcolor{red}{$\downarrow$  0.079} & \textcolor{red}{$\downarrow$ 0.31 }\\
        \cline{2-7}
        \cline{2-7}
        &\multirow{9}{*}{\makecell[l]{15\% missing \\ 15\% noise} } &  & 0.754 & 83.28  & 0.763 & 85.43\\  \cline{3-7}
        && missing 5\%  & \textcolor{red}{$\downarrow$ 0.020 }   & \textcolor{red}{$\downarrow$ 1.50 }  & \textcolor{red}{$\downarrow$  0.018 } & \textcolor{red}{$\downarrow$ 0.96 }\\
        && noise   5\%  & \textcolor{red}{$\downarrow$ 0.001 }   &               $\rightarrow$ 0        & \textcolor{red}{$\downarrow$  0.001 } &               $\rightarrow$ 0      \\
        && missing 10\% & \textcolor{red}{$\downarrow$ 0.031 }   & \textcolor{red}{$\downarrow$ 2.26 }  & \textcolor{red}{$\downarrow$  0.036 } & \textcolor{red}{$\downarrow$ 2.08 }\\
        && noise   10\% & \textcolor{red}{$\downarrow$ 0.002 }   &               $\rightarrow$ 0        &               $\rightarrow$ 0         & \textcolor{red}{$\downarrow$ 0.05 }\\
        && missing 15\% & \textcolor{red}{$\downarrow$ 0.035 }   & \textcolor{red}{$\downarrow$ 2.71 }  & \textcolor{red}{$\downarrow$  0.057 } & \textcolor{red}{$\downarrow$ 3.62 }\\
        && noise   15\% & \textcolor{red}{$\downarrow$ 0.003 }   & \textcolor{red}{$\downarrow$ 0.14 }  & \textcolor{red}{$\downarrow$  0.001 } &              $\rightarrow$ 0       \\
        && missing 30\% & \textcolor{red}{$\downarrow$ 0.119 }   & \textcolor{red}{$\downarrow$ 7.28 }  & \textcolor{red}{$\downarrow$  0.124 } & \textcolor{red}{$\downarrow$ 7.88 } \\ 
        && noise   30\% & \textcolor{red}{$\downarrow$ 0.046 }   & \textcolor{red}{$\downarrow$ 0.16 }  & \textcolor{red}{$\downarrow$  0.003 } & \textcolor{red}{$\downarrow$ 0.23 }\\
        \hline
    \end{tabular}
     }
    \caption{BBFN Robust Training. Results are averaged over three random runs.}
    \label{tab: bbfn_robust}
\end{table*}
        
\begin{table*}[t]
    \centering
     \small
     \resizebox{0.7\textwidth}{!}{
    \begin{tabular}{ |l|l|l|r|r|r|r|  }
        \hline
         &  \multirow{2}{*}{\makecell[l]{\textbf{Robust }\\\textbf{Training}}}& \multirow{2}{*}{ \textbf{Diagnostic} }&\multicolumn{2}{c|}{\textbf{MOSI}} & \multicolumn{2}{c|}{\textbf{MOSEI}}\\
         & & &  Corr & F1 &  Corr & F1\\
        \hline
        \hline
        \multirow{36}{*}{\rotatebox[origin=c]{90}{Self-MM}} &\multirow{9}{*}{-} &   & 0.794 & 85.61 & 0.759 & 84.62\\  \cline{3-7}
        && missing 5\%  & \textcolor{red}{$\downarrow$ 0.023 }   & \textcolor{red}{$\downarrow$ 1.93 }  & \textcolor{red}{$\downarrow$  0.018 } & \textcolor{red}{$\downarrow$ 1.99 }\\
        && noise   5\%  & \textcolor{red}{$\downarrow$ 0.009 }   & \textcolor{red}{$\downarrow$ 0.91 }  & \textcolor{red}{$\downarrow$  0.022 } & \textcolor{red}{$\downarrow$ 1.32 }\\
        && missing 10\% & \textcolor{red}{$\downarrow$ 0.046 }   & \textcolor{red}{$\downarrow$ 3.57 }  & \textcolor{red}{$\downarrow$  0.040 } & \textcolor{red}{$\downarrow$ 3.51 }\\
        && noise   10\% & \textcolor{red}{$\downarrow$ 0.039 }   & \textcolor{red}{$\downarrow$ 2.43 }  & \textcolor{red}{$\downarrow$  0.058 } & \textcolor{red}{$\downarrow$ 2.91 }\\
        && missing 15\% & \textcolor{red}{$\downarrow$ 0.051 }   & \textcolor{red}{$\downarrow$ 4.77 }  & \textcolor{red}{$\downarrow$  0.056 } & \textcolor{red}{$\downarrow$ 4.51 }\\
        && noise   15\% & \textcolor{red}{$\downarrow$ 0.050 }   & \textcolor{red}{$\downarrow$ 2.71 }  & \textcolor{red}{$\downarrow$  0.069 } & \textcolor{red}{$\downarrow$ 3.65 }\\
        && missing 30\% & \textcolor{red}{$\downarrow$ 0.099 }   & \textcolor{red}{$\downarrow$ 11.74}  & \textcolor{red}{$\downarrow$  0.126 } & \textcolor{red}{$\downarrow$ 9.04 }\\
        && noise   30\% & \textcolor{red}{$\downarrow$ 0.154 }   & \textcolor{red}{$\downarrow$ 8.35 }  & \textcolor{red}{$\downarrow$  0.172 } & \textcolor{red}{$\downarrow$ 9.48 }\\
        \cline{2-7}
        \cline{2-7}
        &\multirow{9}{*}{\makecell[l]{5\% missing \\ 5\% noise}} &    & 0.798 & 83.97 & 0.789 & 0.837 \\  \cline{3-7}
        && missing 5\%  & \textcolor{red}{$\downarrow$ 0.022 }  & \textcolor{red}{$\downarrow$  1.73}  & \textcolor{red}{$\downarrow$  0.021 } & \textcolor{red}{$\downarrow$ 1.83 }\\
        && noise   5\%  & \textcolor{red}{$\downarrow$ 0.002 }  &               $\rightarrow$ 0        & \textcolor{red}{$\downarrow$  0.002 } & \textcolor{red}{$\downarrow$ 0.05 }\\
        && missing 10\% & \textcolor{red}{$\downarrow$ 0.046 }  & \textcolor{red}{$\downarrow$ 3.32 }  & \textcolor{red}{$\downarrow$  0.046 } & \textcolor{red}{$\downarrow$ 3.28 }\\
        && noise   10\% & \textcolor{red}{$\downarrow$ 0.012 }  & \textcolor{red}{$\downarrow$ 0.14 }  & \textcolor{red}{$\downarrow$  0.011 } & \textcolor{red}{$\downarrow$ 0.44 }\\
        && missing 15\% & \textcolor{red}{$\downarrow$ 0.050 }  & \textcolor{red}{$\downarrow$ 4.63 }  & \textcolor{red}{$\downarrow$  0.053 } & \textcolor{red}{$\downarrow$ 4.60 }\\
        && noise   15\% & \textcolor{red}{$\downarrow$ 0.018 }  & \textcolor{red}{$\downarrow$ 0.09 }  & \textcolor{red}{$\downarrow$  0.018 } & \textcolor{red}{$\downarrow$ 0.51 }\\
        && missing 30\% & \textcolor{red}{$\downarrow$ 0.119 }  & \textcolor{red}{$\downarrow$ 9.57 }  & \textcolor{red}{$\downarrow$  0.124 } & \textcolor{red}{$\downarrow$ 9.77 }\\
        && noise   30\% & \textcolor{red}{$\downarrow$ 0.046 }  & \textcolor{red}{$\downarrow$ 0.33 }  & \textcolor{red}{$\downarrow$  0.043 } & \textcolor{red}{$\downarrow$ 1.04 }\\
        \cline{2-7}
        \cline{2-7}
        &\multirow{9}{*}{\makecell[l]{10\% missing \\ 10\% noise}} &   & 0.789 & 83.67  & 0.764 & 0.849 \\  \cline{3-7}
        && missing 5\%  & \textcolor{red}{$\downarrow$ 0.021 }  & \textcolor{red}{$\downarrow$ 1.83 }  & \textcolor{red}{$\downarrow$  0.017 } & \textcolor{red}{$\downarrow$ 0.63 }\\
        && noise   5\%  & \textcolor{red}{$\downarrow$ 0.002 }  & \textcolor{red}{$\downarrow$ 0.05 }  & \textcolor{red}{$\downarrow$  0.005 } & \textcolor{red}{$\downarrow$ 0.03 }\\
        && missing 10\% & \textcolor{red}{$\downarrow$ 0.046 }  & \textcolor{red}{$\downarrow$ 3.28 }  & \textcolor{red}{$\downarrow$  0.038 } & \textcolor{red}{$\downarrow$ 1.93 }\\
        && noise   10\% & \textcolor{red}{$\downarrow$ 0.011 }  & \textcolor{red}{$\downarrow$ 0.44 }  & \textcolor{red}{$\downarrow$  0.011 } & \textcolor{red}{$\downarrow$ 0.15 }\\
        && missing 15\% & \textcolor{red}{$\downarrow$ 0.053 }  & \textcolor{red}{$\downarrow$ 4.60 }  & \textcolor{red}{$\downarrow$  0.057 } & \textcolor{red}{$\downarrow$ 3.16 }\\
        && noise   15\% & \textcolor{red}{$\downarrow$ 0.018 }  & \textcolor{red}{$\downarrow$ 0.51 }  & \textcolor{red}{$\downarrow$  0.019 } & \textcolor{red}{$\downarrow$ 0.13 }\\
        && missing 30\% & \textcolor{red}{$\downarrow$ 0.124 }  & \textcolor{red}{$\downarrow$ 9.77 }  & \textcolor{red}{$\downarrow$  0.126 } & \textcolor{red}{$\downarrow$ 7.42 }\\
        && noise   30\% & \textcolor{red}{$\downarrow$ 0.043 }  & \textcolor{red}{$\downarrow$ 1.04 }  & \textcolor{red}{$\downarrow$  0.056 } & \textcolor{red}{$\downarrow$ 1.01 }\\
        \cline{2-7}
        \cline{2-7}
        &\multirow{9}{*}{\makecell[l]{15\% missing \\ 15\% noise} } &  & 0.790 & 84.73 & 0.754 & 84.67\\  \cline{3-7}
        && missing 5\%  & \textcolor{red}{$\downarrow$ 0.022 }  & \textcolor{red}{$\downarrow$ 1.91 }  & \textcolor{red}{$\downarrow$  0.017 } & \textcolor{red}{$\downarrow$ 0.60 }\\
        && noise   5\%  & \textcolor{red}{$\downarrow$ 0.003 }  &              $\rightarrow$ 0         & \textcolor{red}{$\downarrow$  0.004 } & \textcolor{red}{$\downarrow$ 0.01 }\\
        && missing 10\% & \textcolor{red}{$\downarrow$ 0.046 }  & \textcolor{red}{$\downarrow$ 3.39 }  & \textcolor{red}{$\downarrow$  0.038 } & \textcolor{red}{$\downarrow$ 1.91 }\\
        && noise   10\% & \textcolor{red}{$\downarrow$ 0.010 }  & \textcolor{red}{$\downarrow$ 0.44 }  & \textcolor{red}{$\downarrow$  0.011 } & \textcolor{red}{$\downarrow$ 0.14 }\\
        && missing 15\% & \textcolor{red}{$\downarrow$ 0.049 }  & \textcolor{red}{$\downarrow$ 4.58 }  & \textcolor{red}{$\downarrow$  0.055 } & \textcolor{red}{$\downarrow$ 3.08 }\\
        && noise   15\% & \textcolor{red}{$\downarrow$ 0.018 }  & \textcolor{red}{$\downarrow$ 0.28 }  & \textcolor{red}{$\downarrow$  0.016 } & \textcolor{red}{$\downarrow$ 0.16 }\\
        && missing 30\% & \textcolor{red}{$\downarrow$ 0.120 }  & \textcolor{red}{$\downarrow$ 9.66 }  & \textcolor{red}{$\downarrow$  0.122 } & \textcolor{red}{$\downarrow$ 6.86 }\\ 
        && noise   30\% & \textcolor{red}{$\downarrow$ 0.041 }  & \textcolor{red}{$\downarrow$ 0.58 }  & \textcolor{red}{$\downarrow$  0.051 } & \textcolor{red}{$\downarrow$ 1.02 }\\
        \hline
    \end{tabular}
     }
    \caption{Self-MM Robust Training. Results are averaged over three random runs.}
    \label{tab: self_mm_robust}
\end{table*}
        
\begin{table*}[t]
    \centering
     \small
     \resizebox{0.7\textwidth}{!}{
    \begin{tabular}{ |l|l|l|r|r|r|r|  }
        \hline
         &  \multirow{2}{*}{\makecell[l]{\textbf{Robust }\\\textbf{Training}}}& \multirow{2}{*}{ \textbf{Diagnostic} }&\multicolumn{2}{c|}{\textbf{MOSI}} & \multicolumn{2}{c|}{\textbf{MOSEI}}\\
         & & &  Corr & F1 &  Corr & F1\\
        \hline
        \hline
        \multirow{36}{*}{\rotatebox[origin=c]{90}{MMIM}} &\multirow{9}{*}{-} &   & 0.796 & 86.02 & 0.758 & 84.89\\  \cline{3-7}
        && missing 5\%  & \textcolor{red}{$\downarrow$ 0.028 }  & \textcolor{red}{$\downarrow$ 1.34 }  & \textcolor{red}{$\downarrow$  0.016 } & \textcolor{red}{$\downarrow$ 0.93 }\\
        && noise   5\%  & \textcolor{red}{$\downarrow$ 0.035 }  & \textcolor{red}{$\downarrow$ 1.52 }  & \textcolor{red}{$\downarrow$  0.031 } & \textcolor{red}{$\downarrow$ 1.71 }\\
        && missing 10\% & \textcolor{red}{$\downarrow$ 0.067 }  & \textcolor{red}{$\downarrow$ 4.16 }  & \textcolor{red}{$\downarrow$  0.034 } & \textcolor{red}{$\downarrow$ 2.43 }\\
        && noise   10\% & \textcolor{red}{$\downarrow$ 0.058 }  & \textcolor{red}{$\downarrow$ 2.10 }  & \textcolor{red}{$\downarrow$  0.070 } & \textcolor{red}{$\downarrow$ 2.64 }\\
        && missing 15\% & \textcolor{red}{$\downarrow$ 0.056 }  & \textcolor{red}{$\downarrow$ 2.78 }  & \textcolor{red}{$\downarrow$  0.056 } & \textcolor{red}{$\downarrow$ 4.13 }\\
        && noise   15\% & \textcolor{red}{$\downarrow$ 0.078 }  & \textcolor{red}{$\downarrow$ 4.65 }  & \textcolor{red}{$\downarrow$  0.094 } & \textcolor{red}{$\downarrow$ 4.67 }\\
        && missing 30\% & \textcolor{red}{$\downarrow$ 0.117 }  & \textcolor{red}{$\downarrow$ 9.37 }  & \textcolor{red}{$\downarrow$  0.122 } & \textcolor{red}{$\downarrow$ 8.15 }\\
        && noise   30\% & \textcolor{red}{$\downarrow$ 0.197 }  & \textcolor{red}{$\downarrow$ 9.55 }  & \textcolor{red}{$\downarrow$  0.191 } & \textcolor{red}{$\downarrow$ 9.18 }\\
        \cline{2-7}
        \cline{2-7}
        &\multirow{9}{*}{\makecell[l]{5\% missing \\ 5\% noise}} &  & 0.797 & 85.13 & 0.755 & 0.836 \\  \cline{3-7}
        && missing 5\%  & \textcolor{red}{$\downarrow$ 0.057 }  & \textcolor{red}{$\downarrow$ 1.19 }  & \textcolor{red}{$\downarrow$  0.016 } & \textcolor{red}{$\downarrow$ 0.62 }\\
        && noise   5\%  & \textcolor{red}{$\downarrow$ 0.021 }  & \textcolor{red}{$\downarrow$ 1.20 }  & \textcolor{red}{$\downarrow$  0.025 } & \textcolor{red}{$\downarrow$ 1.07 }\\
        && missing 10\% & \textcolor{red}{$\downarrow$ 0.035 }  & \textcolor{red}{$\downarrow$ 1.80 }  & \textcolor{red}{$\downarrow$  0.046 } & \textcolor{red}{$\downarrow$ 2.29 }\\
        && noise   10\% & \textcolor{red}{$\downarrow$ 0.075 }  & \textcolor{red}{$\downarrow$ 2.25 }  & \textcolor{red}{$\downarrow$  0.056 } & \textcolor{red}{$\downarrow$ 2.43 }\\
        && missing 15\% & \textcolor{red}{$\downarrow$ 0.057 }  & \textcolor{red}{$\downarrow$ 4.37 }  & \textcolor{red}{$\downarrow$  0.168 } & \textcolor{red}{$\downarrow$ 2.26 }\\
        && noise   15\% & \textcolor{red}{$\downarrow$ 0.072 }  & \textcolor{red}{$\downarrow$ 3.96 }  & \textcolor{red}{$\downarrow$  0.063 } & \textcolor{red}{$\downarrow$ 2.75 }\\
        && missing 30\% & \textcolor{red}{$\downarrow$ 0.117 }  & \textcolor{red}{$\downarrow$ 8.24 }  & \textcolor{red}{$\downarrow$  0.120 } & \textcolor{red}{$\downarrow$ 6.52 }\\
        && noise   30\% & \textcolor{red}{$\downarrow$ 18.75 }  & \textcolor{red}{$\downarrow$ 9.31 }  & \textcolor{red}{$\downarrow$  0.178 } & \textcolor{red}{$\downarrow$ 7.34 }\\
        \cline{2-7}
        \cline{2-7}
        &\multirow{9}{*}{\makecell[l]{10\% missing \\ 10\% noise}} &   & 0.794 & 84.76 & 0.758 & 84.81 \\  \cline{3-7}
        && missing 5\%  & \textcolor{red}{$\downarrow$ 0.020 }  & \textcolor{red}{$\downarrow$ 1.50 }  & \textcolor{red}{$\downarrow$  0.016 } & \textcolor{red}{$\downarrow$ 1.07 }\\
        && noise   5\%  & \textcolor{red}{$\downarrow$ 0.006 }  & \textcolor{red}{$\downarrow$ 0.46 }  & \textcolor{red}{$\downarrow$  0.024 } & \textcolor{red}{$\downarrow$ 1.28 }\\
        && missing 10\% & \textcolor{red}{$\downarrow$ 0.032 }  & \textcolor{red}{$\downarrow$ 2.09 }  & \textcolor{red}{$\downarrow$  0.043 } & \textcolor{red}{$\downarrow$ 2.79 }\\
        && noise   10\% & \textcolor{red}{$\downarrow$ 0.060 }  & \textcolor{red}{$\downarrow$ 2.57 }  & \textcolor{red}{$\downarrow$  0.045 } & \textcolor{red}{$\downarrow$ 2.58 }\\
        && missing 15\% & \textcolor{red}{$\downarrow$ 0.060 }  & \textcolor{red}{$\downarrow$ 3.27 }  & \textcolor{red}{$\downarrow$  0.058 } & \textcolor{red}{$\downarrow$ 3.84 }\\
        && noise   15\% & \textcolor{red}{$\downarrow$ 0.062 }  & \textcolor{red}{$\downarrow$ 3.65 }  & \textcolor{red}{$\downarrow$  0.063 } & \textcolor{red}{$\downarrow$ 3.42 }\\
        && missing 30\% & \textcolor{red}{$\downarrow$ 0.110 }  & \textcolor{red}{$\downarrow$ 8.40 }  & \textcolor{red}{$\downarrow$  0.119 } & \textcolor{red}{$\downarrow$ 7.50 }\\
        && noise   30\% & \textcolor{red}{$\downarrow$ 0.163 }  & \textcolor{red}{$\downarrow$ 9.18 }  & \textcolor{red}{$\downarrow$  0.177 } & \textcolor{red}{$\downarrow$ 8.16 }\\
        \cline{2-7}
        \cline{2-7}
        &\multirow{9}{*}{\makecell[l]{15\% missing \\ 15\% noise} } &  & 0.784 & 84.67 & 0.751 & 83.15\\  \cline{3-7}
        && missing 5\%  & \textcolor{red}{$\downarrow$ 0.014 }  & \textcolor{red}{$\downarrow$ 1.79 }  & \textcolor{red}{$\downarrow$  0.020 } & \textcolor{red}{$\downarrow$ 0.89 }\\
        && noise   5\%  & \textcolor{red}{$\downarrow$ 0.028 }  & \textcolor{red}{$\downarrow$ 1.04 }  & \textcolor{red}{$\downarrow$  0.016 } & \textcolor{red}{$\downarrow$ 0.69 }\\
        && missing 10\% & \textcolor{red}{$\downarrow$ 0.038 }  & \textcolor{red}{$\downarrow$ 2.18 }  & \textcolor{red}{$\downarrow$  0.030 } & \textcolor{red}{$\downarrow$ 2.24 }\\
        && noise   10\% & \textcolor{red}{$\downarrow$ 0.050 }  & \textcolor{red}{$\downarrow$ 1.30 }  & \textcolor{red}{$\downarrow$  0.028 } & \textcolor{red}{$\downarrow$ 0.62 }\\
        && missing 15\% & \textcolor{red}{$\downarrow$ 0.061 }  & \textcolor{red}{$\downarrow$ 4.91 }  & \textcolor{red}{$\downarrow$  0.059 } & \textcolor{red}{$\downarrow$ 2.77 }\\
        && noise   15\% & \textcolor{red}{$\downarrow$ 0.071 }  & \textcolor{red}{$\downarrow$ 4.08 }  & \textcolor{red}{$\downarrow$  0.040 } & \textcolor{red}{$\downarrow$ 1.58 }\\
        && missing 30\% & \textcolor{red}{$\downarrow$ 0.146 }  & \textcolor{red}{$\downarrow$ 10.48}  & \textcolor{red}{$\downarrow$  0.115 } & \textcolor{red}{$\downarrow$ 6.62 } \\ 
        && noise   30\% & \textcolor{red}{$\downarrow$ 0.180 }  & \textcolor{red}{$\downarrow$ 8.88 }  & \textcolor{red}{$\downarrow$  0.096 } & \textcolor{red}{$\downarrow$ 4.41 } \\
        \hline
    \end{tabular}
     }
    \caption{MMIM Robust Training. Results are averaged over three random runs.}
    \label{tab: mmim_robust}
\end{table*}
        
\begin{table*}[t]
    \centering
     \small
     \resizebox{0.7\textwidth}{!}{
    \begin{tabular}{ |l|l|l|r|r|r|r|  }
        \hline
         &  \multirow{2}{*}{\makecell[l]{\textbf{Robust }\\\textbf{Training}}}& \multirow{2}{*}{ \textbf{Diagnostic} }&\multicolumn{2}{c|}{\textbf{MOSI}} & \multicolumn{2}{c|}{\textbf{MOSEI}}\\
         & & &  Corr & F1 &  Corr & F1\\
        \hline
        \hline
        \multirow{36}{*}{\rotatebox[origin=c]{90}{MulT}} &\multirow{9}{*}{-} &   & 0.747 & 82.25 & 0.738 & 83.37\\  \cline{3-7}
        && missing 5\%  & \textcolor{red}{$\downarrow$ 0.032 }  & \textcolor{red}{$\downarrow$ 3.76 }  & \textcolor{red}{$\downarrow$ 0.019} & \textcolor{red}{$\downarrow$ 0.84}\\
        && noise   5\%  & \textcolor{red}{$\downarrow$ 0.053 }  & \textcolor{red}{$\downarrow$ 1.67 }  &             $\rightarrow$ 0         &             $\rightarrow$ 0       \\
        && missing 10\% & \textcolor{red}{$\downarrow$ 0.046 }  & \textcolor{red}{$\downarrow$ 3.92 }  & \textcolor{red}{$\downarrow$ 0.031} & \textcolor{red}{$\downarrow$ 1.85}\\
        && noise   10\% & \textcolor{red}{$\downarrow$ 0.069 }  & \textcolor{red}{$\downarrow$ 3.03 }  &             $\rightarrow$ 0         &             $\rightarrow$ 0       \\
        && missing 15\% & \textcolor{red}{$\downarrow$ 0.052 }  & \textcolor{red}{$\downarrow$ 4.84 }  & \textcolor{red}{$\downarrow$ 0.051} & \textcolor{red}{$\downarrow$ 3.29}\\
        && noise   15\% & \textcolor{red}{$\downarrow$ 0.172 }  & \textcolor{red}{$\downarrow$ 5.00 }  & \textcolor{red}{$\downarrow$ 0.152} & \textcolor{red}{$\downarrow$ 4.50}\\
        && missing 30\% & \textcolor{red}{$\downarrow$ 0.113 }  & \textcolor{red}{$\downarrow$ 12.17}  & \textcolor{red}{$\downarrow$ 0.109} & \textcolor{red}{$\downarrow$ 6.60}\\
        && noise   30\% & \textcolor{red}{$\downarrow$ 0.295 }  & \textcolor{red}{$\downarrow$ 8.99 }  & \textcolor{red}{$\downarrow$ 0.263} & \textcolor{red}{$\downarrow$ 7.95}\\
        \cline{2-7}
        \cline{2-7}
        &\multirow{9}{*}{\makecell[l]{5\% missing \\ 5\% noise}} &  & 0.748 & 81.90 & 0.748 & 84.59 \\  \cline{3-7}
        && missing 5\%  & \textcolor{red}{$\downarrow$ 0.031 }  & \textcolor{red}{$\downarrow$ 1.50 }  & \textcolor{red}{$\downarrow$  0.018 } & \textcolor{red}{$\downarrow$ 0.93 }\\
        && noise   5\%  & \textcolor{red}{$\downarrow$ 0.021 }  & \textcolor{red}{$\downarrow$ 1.67 }  &              $\rightarrow$ 0          &         $\rightarrow$ 0    \\
        && missing 10\% & \textcolor{red}{$\downarrow$ 0.043 }  & \textcolor{red}{$\downarrow$ 2.43 }  & \textcolor{red}{$\downarrow$  0.030 } & \textcolor{red}{$\downarrow$ 1.99 }\\
        && noise   10\% & \textcolor{red}{$\downarrow$ 0.021 }  & \textcolor{red}{$\downarrow$ 1.96 }  &              $\rightarrow$ 0          &          $\rightarrow$ 0    \\
        && missing 15\% & \textcolor{red}{$\downarrow$ 0.050 }  & \textcolor{red}{$\downarrow$ 3.36 }  & \textcolor{red}{$\downarrow$  0.050 } & \textcolor{red}{$\downarrow$ 3.51 }\\
        && noise   15\% & \textcolor{red}{$\downarrow$ 0.067 }  & \textcolor{red}{$\downarrow$ 3.30 }  & \textcolor{red}{$\downarrow$  0.173 } & \textcolor{red}{$\downarrow$ 3.97 }\\
        && missing 30\% & \textcolor{red}{$\downarrow$ 0.077 }  & \textcolor{red}{$\downarrow$ 5.12 }  & \textcolor{red}{$\downarrow$  0.074 } & \textcolor{red}{$\downarrow$ 4.90 }\\
        && noise   30\% & \textcolor{red}{$\downarrow$ 0.094 }  & \textcolor{red}{$\downarrow$ 5.17 }  & \textcolor{red}{$\downarrow$  0.198 } & \textcolor{red}{$\downarrow$ 5.50 }\\
        \cline{2-7}
        \cline{2-7}
        &\multirow{9}{*}{\makecell[l]{10\% missing \\ 10\% noise}} &   & 0.741 & 81.31 & 0.746 & 84.13 \\  \cline{3-7}
        && missing 5\%  & \textcolor{red}{$\downarrow$ 0.034 }  & \textcolor{red}{$\downarrow$ 2.57 }  & \textcolor{red}{$\downarrow$  0.019 } & \textcolor{red}{$\downarrow$ 0.96 }\\
        && noise   5\%  & \textcolor{red}{$\downarrow$ 0.012 }  & \textcolor{red}{$\downarrow$ 0.09 }  &              $\rightarrow$ 0          &              $\rightarrow$ 0       \\
        && missing 10\% & \textcolor{red}{$\downarrow$ 0.046 }  & \textcolor{red}{$\downarrow$ 3.06 }  & \textcolor{red}{$\downarrow$  0.030 } & \textcolor{red}{$\downarrow$ 0.019 }\\
        && noise   10\% & \textcolor{red}{$\downarrow$ 0.031 }  & \textcolor{red}{$\downarrow$ 2.42 }  &              $\rightarrow$ 0          &              $\rightarrow$ 0        \\
        && missing 15\% & \textcolor{red}{$\downarrow$ 0.053 }  & \textcolor{red}{$\downarrow$ 5.52 }  & \textcolor{red}{$\downarrow$  0.048 } & \textcolor{red}{$\downarrow$ 3.19 }\\
        && noise   15\% & \textcolor{red}{$\downarrow$ 0.045 }  & \textcolor{red}{$\downarrow$ 4.08 }  & \textcolor{red}{$\downarrow$  0.153 } & \textcolor{red}{$\downarrow$ 4.08 }\\
        && missing 30\% & \textcolor{red}{$\downarrow$ 0.077 }  & \textcolor{red}{$\downarrow$ 6.57 }  & \textcolor{red}{$\downarrow$  0.072 } & \textcolor{red}{$\downarrow$ 4.62 }\\
        && noise   30\% & \textcolor{red}{$\downarrow$ 0.056 }  & \textcolor{red}{$\downarrow$ 4.57 }  & \textcolor{red}{$\downarrow$  0.202 } & \textcolor{red}{$\downarrow$ 5.58 }\\
        \cline{2-7}
        \cline{2-7}
        &\multirow{9}{*}{\makecell[l]{15\% missing \\ 15\% noise} } &  & 0.744 & 82.21 & 0.745 & 83.95\\  \cline{3-7}
        && missing 5\%  & \textcolor{red}{$\downarrow$ 0.034 }  & \textcolor{red}{$\downarrow$ 2.56 }  & \textcolor{red}{$\downarrow$  0.018 } & \textcolor{red}{$\downarrow$ 0.86 }\\
        && noise   5\%  & \textcolor{red}{$\downarrow$ 0.023 }  & \textcolor{red}{$\downarrow$ 1.52 }  &             $\rightarrow$ 0           &             $\rightarrow$ 0        \\
        && missing 10\% & \textcolor{red}{$\downarrow$ 0.047 }  & \textcolor{red}{$\downarrow$ 3.93 }  & \textcolor{red}{$\downarrow$  0.032 } & \textcolor{red}{$\downarrow$ 1.86 }\\
        && noise   10\% & \textcolor{red}{$\downarrow$ 0.039 }  & \textcolor{red}{$\downarrow$ 3.62 }  &             $\rightarrow$ 0           &             $\rightarrow$ 0        \\
        && missing 15\% & \textcolor{red}{$\downarrow$ 0.052 }  & \textcolor{red}{$\downarrow$ 5.15 }  & \textcolor{red}{$\downarrow$  0.051 } & \textcolor{red}{$\downarrow$ 3.27 }\\
        && noise   15\% & \textcolor{red}{$\downarrow$ 0.046 }  & \textcolor{red}{$\downarrow$ 3.62 }  &             $\rightarrow$ 0           & \textcolor{red}{$\downarrow$ 0.08 }\\
        && missing 30\% & \textcolor{red}{$\downarrow$ 0.117 }  & \textcolor{red}{$\downarrow$ 9.63 }  & \textcolor{red}{$\downarrow$ 0.113}   & \textcolor{red}{$\downarrow$  7.03} \\ 
        && noise   30\% & \textcolor{red}{$\downarrow$ 0.068 }  & \textcolor{red}{$\downarrow$ 6.13 }  & \textcolor{red}{$\downarrow$ 0.001}   & \textcolor{darkgreen}{$\uparrow$ 0.02} \\
        \hline
    \end{tabular}
     }
    \caption{MulT Robust Training. Results are averaged over three random runs.}
    \label{tab: mult_robust}
\end{table*}

\end{document}